\documentclass[conference]{IEEEtran}
\IEEEoverridecommandlockouts

\usepackage{cite}
\usepackage{amsmath,amssymb,amsfonts}
\usepackage{algorithmic}
\usepackage{graphicx}
\usepackage{textcomp}
\usepackage{xcolor}
\usepackage{tabularray}
\usepackage{url}
\usepackage{wasysym}
\usepackage{hyperref}
\usepackage[capitalize,noabbrev]{cleveref}
\def\BibTeX{{\rm B\kern-.05em{\sc i\kern-.025em b}\kern-.08em
    T\kern-.1667em\lower.7ex\hbox{E}\kern-.125emX}}
\begin{document}

\title{Mitigating Hallucinations on Object Attributes  using Multiview Images and Negative Instructions \thanks{This research is supported by the
 National Key R\&D Program of China (2023YFC3304902).  The HoOA benchmark is available at \href{here}\url{{https://drive.google.com/file/d/1o4MI3GTSOAGu_EBtcYuCXEz17Mn5rNOR/}}.}
}

\author{
	\IEEEauthorblockN{
		Zhijie Tan\IEEEauthorrefmark{1}$^{,\ddag}$, 
		Yuzhi Li\IEEEauthorrefmark{1}$^{,\ddag}$, 
		Shengwei Meng\IEEEauthorrefmark{2}, 
		Xiang Yuan\IEEEauthorrefmark{1},
            Weiping Li\IEEEauthorrefmark{1}$^{,\clubsuit}$,
		Tong Mo\IEEEauthorrefmark{1},
            Bingce Wang\IEEEauthorrefmark{1},
            Xu Chu \IEEEauthorrefmark{1}} 
	\IEEEauthorblockA{\IEEEauthorrefmark{1}School of Software and Microelectronics, Peking University, Beijing, China }
	\IEEEauthorblockA{\IEEEauthorrefmark{2}School of Computer Science, Beijing University of Posts and Telecommunications, Beijing, China}
 	\IEEEauthorblockA{\IEEEauthorrefmark{3}Both authors contributed equally. $^\clubsuit$Corresponding author.}
        \IEEEauthorblockA{EMail: besttangent@stu.pku.edu.cn, nidhogg\_lyz@stu.pku.edu.cn, mengshengwei@bupt.edu.cn, \\ xiangyuan@stu.pku.edu.cn, wpli@ss.pku.edu.cn, motong@ss.pku.edu.cn, \{wangbingce, chuxu\}@stu.pku.edu.cn}
} 

\maketitle

\begin{abstract}
Current popular Large Vision-Language Models (LVLMs) are suffering from Hallucinations on Object Attributes (HoOA), leading to incorrect determination of fine-grained attributes in the input images. Leveraging significant advancements in 3D generation from a single image, this paper proposes a novel method to mitigate HoOA in LVLMs. This method utilizes multiview images sampled from generated 3D representations as visual prompts for LVLMs, thereby providing more visual information from other viewpoints. Furthermore, we observe the input order of multiple multiview images significantly affects the performance of LVLMs. Consequently, we have devised Multiview Image Augmented VLM (MIAVLM), incorporating a Multiview Attributes Perceiver (MAP) submodule capable of simultaneously eliminating the influence of input image order and aligning visual information from multiview images with Large Language Models (LLMs). Besides, we designed and employed negative instructions to mitigate LVLMs' bias towards ``Yes" responses. Comprehensive experiments demonstrate the effectiveness of our method.
\end{abstract}


\section{Introduction}

\begin{figure}[tbp]
\begin{center}

\centerline{\includegraphics[width=0.9\columnwidth]{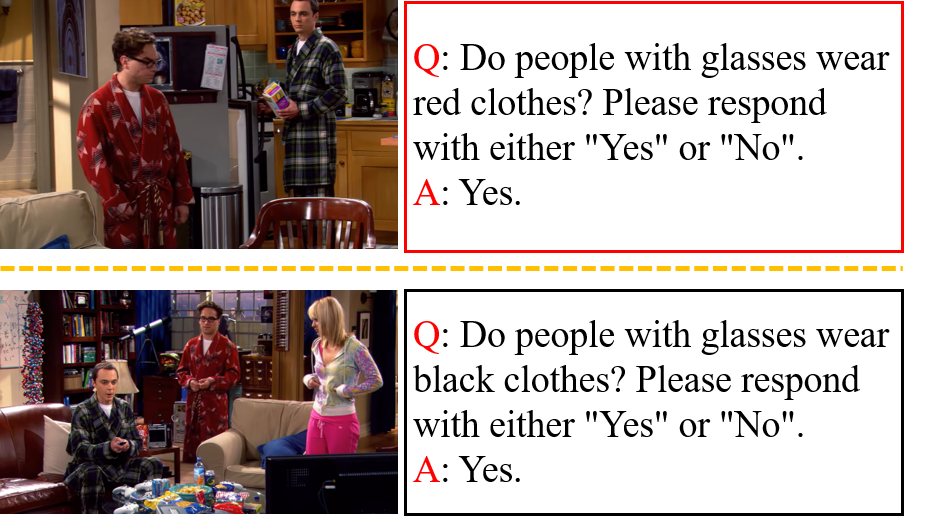}}
\vspace{-1em}
\caption{
Illustration of the HoOE Problem.}
\label{fig: intro}
\end{center}
\vspace{-4em}
\end{figure}

Current popular Large Vision-Language Models (LVLMs) \cite{chen2024large,xu2024llava,zhang2024vision,chen2023vlp,fadeeva2024representing,tan2023building} are suffering from hallucinations \cite{liu2024survey,li2023evaluating}. These hallucinations manifest as inconsistencies between the textual responses generated by LVLMs and the semantic content of input images \cite{rawte2023survey}. Specifically, these hallucinations can be categorized into three \textbf{\textit{types}} \cite{liu2024survey}: \textbf{\textit{a.) Hallucination on Object Existence (HoOE)}}, wherein errors occur in judgments regarding the presence of objects, such as when non-existent objects are included in the descriptions generated by LVLMs; \textbf{\textit{b.) Hallucination on Object Attributes (HoOA) }}, wherein errors arise in describing the attributes of objects, including shape and color attributes, as exemplified by LVLMs describing a red apple as green; and \textbf{\textit{c.) Hallucination on Object Relationships (HoOR) }}, wherein errors occur in describing relationships between different objects, such as describing a person in front of a sofa as being behind it \cite{liu2024survey}. Notably, benchmarks designed for assessing LVLMs' hallucinations, such as M-HalDetect \cite{gunjal2024detecting}, MMHal-Bench \cite{sun2023aligning}, and AMBER \cite{wang2023llm}, include multiple objects that exhibit issues related to existence, attributes, and relationships simultaneously. Consequently, these three types of problems are strongly coupled in current benchmarks, posing significant challenges for analyzing their individual causes. For instance, in addressing the HoOA, the presence of multiple objects introduces hallucinations related to HoOE and HoOR, thereby complicating the analysis. More specifically, as illustrated in Figure \ref{fig: intro}, LLaVA-1.5 \cite{liu2023improved} provides correct answers to questions within the red box above the dashed line. However, in the image below the dashed line, LLaVA-1.5 determines that there is a person wearing glasses and dressed in black. This constitutes a HoOE problem. However, such hallucinations might be caused by the LVLMs failing to correctly understand the ``black" attribute, which corresponds to a HoOA problem. In such complex test scenarios, it is challenging to decouple these different types of hallucinations and address them separately, making it difficult to accurately assess the true capabilities of LVLMs.

Therefore, it is necessary to design individual benchmarks for each type of hallucination that can exclude interference from other hallucinations. This article demonstrates how to utilize face captioning as a foundational task to design a benchmark for the HoOA problem. Face captioning is a crucial multimodal task widely employed in downstream applications such as facial recognition \cite{li2022talk2face} and text-to-face applications \cite{sun2022anyface}. The CelebAText-HQ \cite{sun2021multi} dataset, manually annotated with facial attributes, provides detailed descriptions for each face, including shape, color, and other facial attributes. CelebAText-HQ exclusively offers detailed descriptions for individual objects (faces), thereby allowing us to design a benchmark that excludes issues related to HoOE and HoOR, facilitating a more accurate evaluation of the HoOA problem. In constructing this benchmark, we employ common techniques used in standard evaluation metrics, such as POPE \cite{li2023evaluating}, CIEM \cite{hu2023ciem}, and NOPE \cite{lovenia2023negative}, to transform the generative task into a discriminative task. Each manually annotated description is converted into a question posed to LVLMs, with occurrences of “Yes” responses tallied to calculate \textit{accuracy}. 
It is noteworthy that all converted questions yield “Yes” answers. However, as mentioned in previous studies \cite{liu2023mitigating,hu2023ciem}, there exists a tendency in current LVLMs to favor “Yes” responses disproportionately. Consequently, to assess whether LVLMs recognize the attributes of the images, we designed questions for which the answer is ``No" for the same image. For clarity, we term questions with correct ``Yes" answers as positive questions, and those with ``No" answers as negative questions. Ultimately, we observed a near-opposite performance of LVLMs on positive and negative questions.

Training high-quality LVLMs requires addressing both training data and model design aspects. Therefore, we analyze potential causes of HoOA from both data and model perspectives. From the data perspective, the emergence of the HoOA problem can be ascribed to two \textit{\textbf{causes}}: \textit{\textbf{a.)}} Insufficient information in single images to enable LVLMs to generate correct responses. \textit{\textbf{b.)}} Popular LVLMs often undergo instruction tuning with a high proportion of positive visual instructions.

In response to \textit{\textbf{cause a.)}}, prior studies have found that introducing richer image descriptions \cite{gunjal2024detecting} or spatial information \cite{you2023ferret} can effectively mitigate hallucinations in LVLMs. An intuitive approach is to introduce additional depth maps to significantly improve HoOR problems. Consider the relationship between a person and a sofa: introducing a depth map as additional information can effectively resolve HoOR problems based on the different depths of the sofa and the person. However, when considering the HoOA problem, solely utilizing semantic segmentation or depth maps from the current viewpoint would evidently overlook fine-grained attribute information from other viewpoints. This loss of attribute information leads to two results. Firstly, certain fine-grained details from the current viewpoint may be incomplete. In cases where questions are posed about these potentially incomplete details, the possibility of LVLMs producing hallucinations exists. Secondly, fine-grained attribute information from other viewpoints is almost certainly incomplete. When questions are posed about these inherently incomplete details, LVLMs are highly likely to produce hallucinations. Therefore, generating 3D representations for current objects can effectively mitigate such HoOA problems. Benefiting from the considerable advancements in generating 3D representations from single images, images from other viewpoints can be sampled from the 3D representation \cite{voleti2024sv3d,liu2024one,tang2023make}. From the perspective of visual prompt learning \cite{lu2023can, wang2024hierarchical, kirillov2023segment}, these sampled images can be regarded as visual prompts. These visual prompts provide more visual information for the same attributes, thereby enhancing the robustness of LVLMs' responses. As for \textit{\textbf{cause b.)}}, prior works have introduced additional negative visual instructions during instruction tuning to enhance overall performance \cite{liu2023mitigating,you2023ferret}. Drawing inspiration from these approaches, we adopt the aforementioned negative questions as negative instructions to teach the LVLMs to respond “No” to HoOA problems. 

From the model perspective, we have observed that \textit{\textbf{cause c):}} Directly inputting multiple images to LVLMs may lead to HoOA problems. Similar to  LLMs' sensitivity to the order of multiple prompts \cite{tan2024order,tan2024llm,lu2021fantastically}, LVLMs also exhibit sensitivity to the order of multiview images. For \textit{\textbf{causes c)}}, we designed a submodule named \textbf{M}ultiview \textbf{A}ttributes \textbf{P}erceiver (\textbf{MAP}) and integrated it into a model called \textbf{M}ultiview \textbf{I}mage \textbf{A}ugmented \textbf{VLM} (\textbf{MIAVLM}) to align the multiview images with the LLM and mitigate the impact of the input order.

In summary, the contributions of this paper are as follows:

\textbf{1.}To ascertain the presence of the HoOA problem while eliminating interference from HoOE and HoOR problems, we propose the HoOA benchmark. \textbf{2.} To mitigate the HoOA problem, we propose utilizing multiview images of current objects as visual prompts. Furthermore, we design a novel network module called \textbf{MIAVLM}, integrating a \textbf{MAP} submodule capable of eliminating the influence of input image order and aligning visual information from multiview images with LLMs. Additionally, we designed and employed negative instructions to mitigate LVLMs' bias towards "Yes" responses. \textbf{3.} To validate the effectiveness of our algorithms, we conducted comprehensive experiments on the HoOA benchmark.

\begin{figure}[htbp]
\begin{center}
\vspace{-1.3em}
\centerline{\includegraphics[width=0.9\columnwidth]{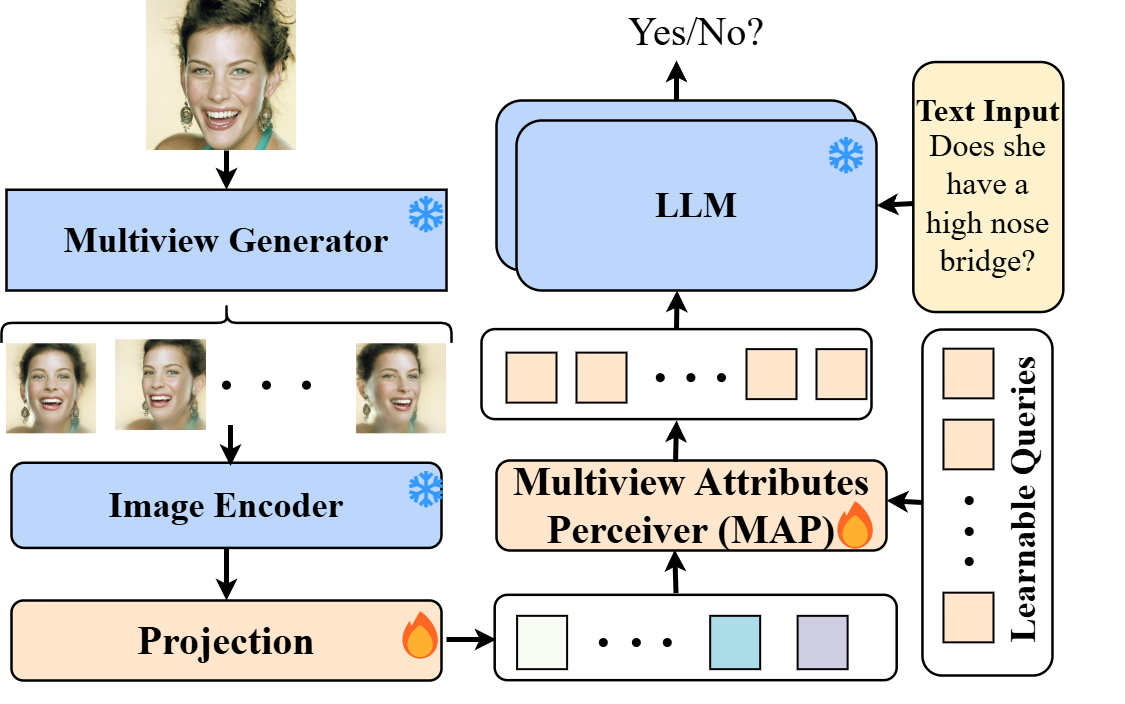}}
\vspace{-1.5em}
\caption{An overview of the MIAVLM model. 
Frozen parts are blue and marked with a snowflake while trainable parts are red and marked with  a flame.}
\label{fig: overview}
\end{center}
\end{figure}
\vspace{-2.5em}

\section{Method}
We propose \textbf{M}ultiview \textbf{I}mage \textbf{A}ugmented \textbf{V}ision-\textbf{V}anguage \textbf{M}odel (MIAVLM), a model for generating more comprehensive and reliable results from multiple inputs. 

\vspace{-1em}

\subsection{Model Architecture}
The overview of the MIAVLM model is shown in \cref{fig: overview}, in which we propose a \textbf{M}ultiview \textbf{A}ttributes \textbf{P}erceiver (\textbf{MAP})  to bridge the gap between a frozen image encoder and a frozen LLM (Flan-T5-large \cite{flat5}). Firstly, the input image is processed by a Multiview Generator (HFGI3D\cite{xie2023high}) to generate multiview images of the input. Secondly, the image encoder (ViT-L/16 \cite{dosovitskiy2020vit}) encodes multiview images into dense embeddings and passes the projected embeddings through the MAP to get the aggregated representation of all the inputs.  Finally, the frozen LLM then accepts the output from the MAP and produces the final text outputs. The inner structure of the MAP is shown in \cref{fig: MAP}, including a Visual Extractor and a Multihead Sampler. The details will be further discussed.

\vspace{-0.5em}

\subsection{Visual Extractor}

Following the Vanilla transformer, we design the Visual Extractor as a pile of transformer decoder blocks (6 blocks). In cross attention, the input soft prompts are regarded as the queries, and the image embeddings are regarded as keys and values to inject the visual information into the soft prompts.

Engaging soft prompts in cross-attention with image embeddings encourages the prompt to interact with vision information and better extract the information for downstream tasks. Since there are multiple input image embeddings, the Visual Extractor performs cross-attention on the soft prompts with each image embedding separately. Formally, we denote the multiple input image embeddings as $E = \{e_1, e_2, ..., e_n\}$, where $e_i$ is the $i$-th input embedding. We denote the soft prompts as $P$ containing $l$ soft tokens, $P \in R^{l \times d}$, where $d$ is the LLM's model dimension ($l$=32 and $d$=1024). 
Denote the output from the Visual Extractor as $O_{VE}$, the mapping matrix as $W_Q$, $W_K$ and $W_V$, then $O_{VE}$ can be computed as follows:

{\small
\begin{align}
    &O_{VE} = \{softmax(\frac{(PW_Q)(e_iW_K)^T}{\sqrt{d}}){e_iW_V} | e_i \in E\} \notag \\ 
    &E = \{e_1, e_2, ..., e_n\}; W_Q, W_K, W_V \in R^{d \times d} \label{VE}
\end{align}
}
In \cref{VE}, $P$ denotes the soft prompts and $e_i$ is the image embedding of the $i$-th input. Noticing that different input has different contributions to the final output, the outputs in $O_{VE}$ are weighted and summed according to the weights computed by the Multihead Sampler. Besides, we compute each cross-attention output in parallel and separately instead of using the previous output as the next query in \cref{VE}. This is because in most conditions the input images have no order and we're supposed to compute their relation to the soft prompts separately.
\vspace{-1em}

\begin{figure}[tbp]
\begin{center}
\centerline{\includegraphics[width=1.0\columnwidth]{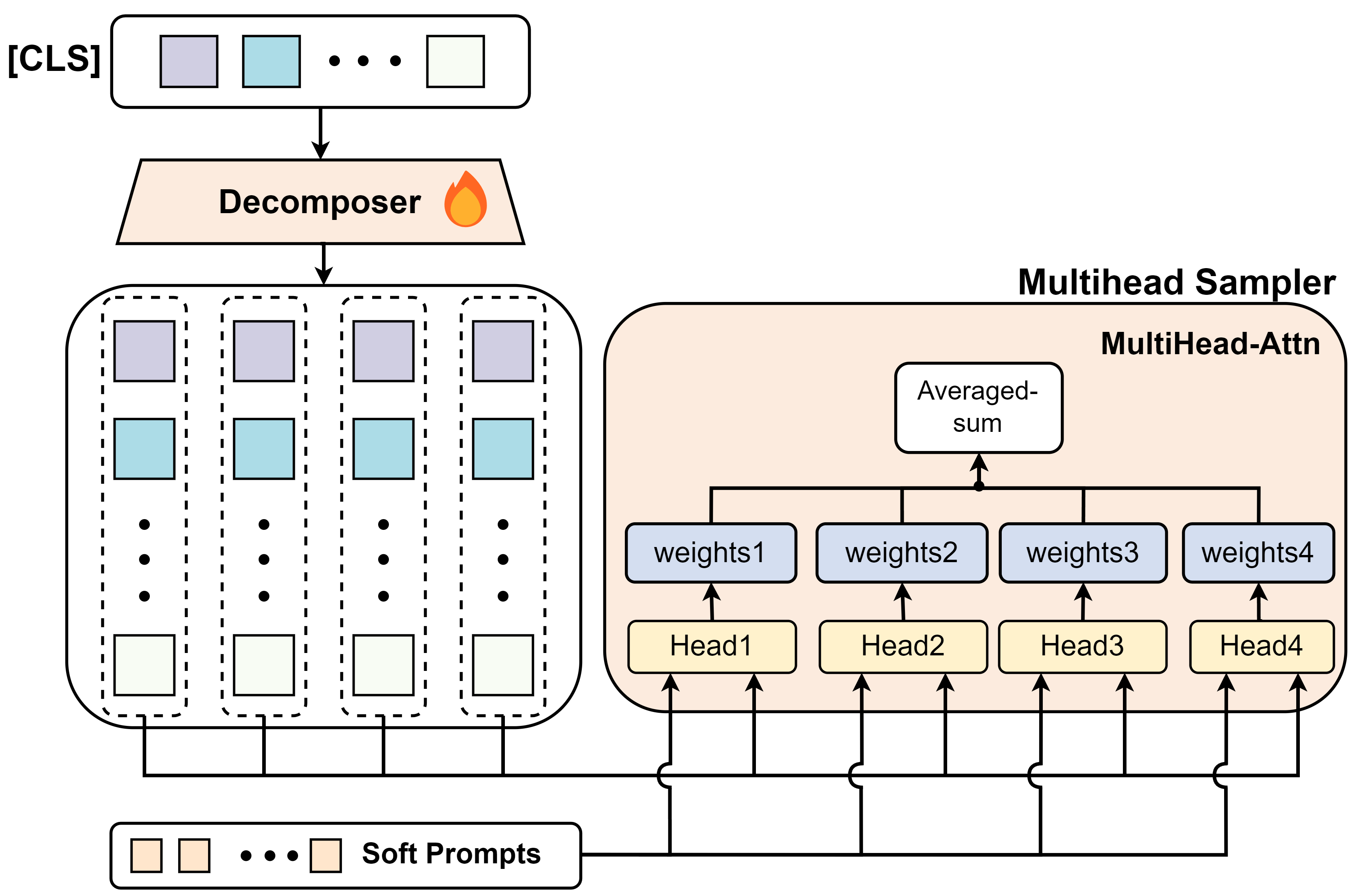}}
\vspace{-1em}
\caption{An overview of the Multihead Sampler. 
}
\label{fig: ms}
\end{center}
\vspace{-4em}
\end{figure}

\subsection{Multihead Sampler}
The Multihead Sampler is used for computing weights for the weighted sum of the Visual Extractor's outputs $O_{VE}$ in \cref{VE}. To further decompose the visual information in the input image embeddings, a Decomposer consisting of a 2-layer MLP is used to map the [CLS] token of the input embedding into $m$ ($m=4$) extra tokens, and the same number of attention heads are applied to compute the attention weights over each decomposed token and the soft prompts. This design aims to introduce multiple experts in the form of attention heads to focus on different features in the inputs.

As shown in \cref{fig: ms}, the soft prompts serve as the queries and the decomposed image embeddings serve as the keys to compute the attention weights. Note that only the attention weights are computed and the means over the query's dimension are taken as the output of each head. Denote $e_i^j$ as the $j$-th decomposed token of the $i$-th input embedding, each head's output $weights_j$ can be written as follows:
\vspace{-1em}

{\small
\begin{flalign}
& score_j = head_j(P, E^j) ; P \in R^{l \times d}, E^j = [e_1^j, e_2^j, ..., e_n^j] \in R^{n \times d} \notag \\
& weights_j = mean(score_j) ; socre_j \in R^{l \times n}; \quad weights_j \in R^{n} \label{eq:2}
\end{flalign}
}
In \cref{eq:2}, $d$ is the model dimension, $P$ is the soft prompts and $head_j$ means the computation of attention score over $P$ and $E_j$. $l$ is the number of tokens in soft prompts. The $mean$ operation takes the averaged sum over the number of tokens in $P$. The averaged sum of weights from each head is taken as the output of the Multihead Sampler:
\vspace{-1em}

{\small
\begin{align}
& w_{MS} =  \frac{1}{m} \sum \limits_{j=1}^{m} weights_j; \quad w_{MS} \in R^{n} \label{eq:3}
\end{align}
}
In \cref{eq:3}, $m$ ($m=4$) is the number of decomposed extra tokens and the number of attention heads in MS.

MS aims to further capture the fine-grained visual features in the input image embeddings by applying different attention heads for different decomposed embeddings of the inputs.
\vspace{-2em}

\begin{figure}[htbp]
\begin{center}
\centerline{\includegraphics[width=\columnwidth]{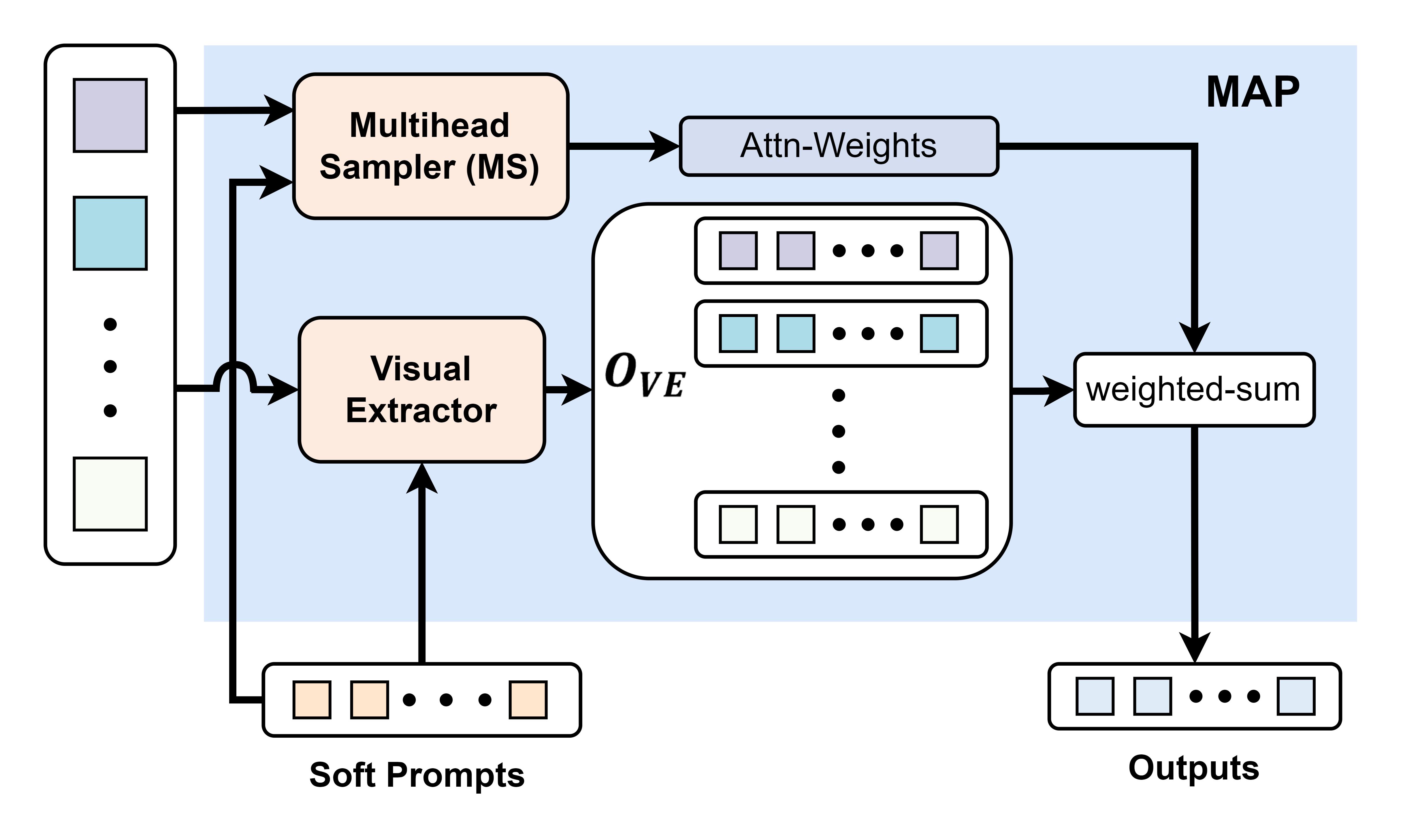}}
\vspace{-1em}
\caption{The structure of Multiview Attributes Perceiver. 
}
\label{fig: MAP}
\end{center}
\vspace{-3.5em}
\end{figure}


\begin{table*}
\centering
\caption{Main results on our HoOA benchmark. Bold is the best.}
\vspace{-1em}
\label{tab:lvlms}
\begin{tblr}{
  width = \linewidth,
  colspec = {Q[215]Q[129]Q[113]Q[125]Q[135]Q[137]Q[104]},
  row{even} = {c},
  row{3} = {c},
  row{5} = {c},
  row{7} = {c},
  row{9} = {c},
  cell{1}{2} = {c},
  cell{1}{5} = {c},
  cell{1}{6} = {c},
  cell{1}{7} = {c},
  cell{2}{1} = {r=3}{},
  cell{5}{1} = {r=6}{},
  hline{1-2,5,11} = {-}{},
}
                                 & Model         & Parameters (B) & Inferece Time (s)              & Positive Accuracy              & Negative Accuracy~             & HoOA metric                     \\
origin image/ 9in1 image         & BLIP3         & 3.9            & 6.021/ 6.149                   & 0.823/ 0.831                   & 0.312/ 0.267                   & 0.568/ 0.549                    \\
                                 & OPERA         & 7.0            & 41.572/ 41.584                 & 0.934/ 0.937                   & 0.152/ 0.107                   & 0.543/ 0.522                    \\
                                 & LLaVA-UHD     & 7.0            & 2.545/ 2.736                   & 0.933/ 0.921                   & 0.157/ 0.113                   & 0.545/ 0.517                    \\
origin image/ 9 multiview images & OpenFlamingo1 & 2.5            & 0.807/ 1.332                        & 0.734/ 0.768                   & 0.385/ 0.397                   & 0.560/ 0.582                    \\
                                 & OpenFlamingo2 & 2.5            & 0.881/ 1.457                        & \textbf{0.963}/ \textbf{0.960} & 0.210/ 0.223                   & 0.606/ 0.591                    \\
                                 & OpenFlamingo3 & 4.9            & 1.112/ 3.152                        & 0.740/ 0.761                   & 0.472/ 0.486                   & 0.606/ 0.623                    \\
                                 & OpenFlamingo4 & 4.9            & 1.247/ 1.793                        & 0.624/ 0.632                   & 0.483/ 0.501                   & 0.553/ 0.565                    \\
                                 & Idefics2     & 8.0            & 1.294/ 6.482                   & 0.847/ 0.852                   & 0.421/ 0.432                   & 0.634/ 0.642                    \\
                                 & MIAVLM (Ours) & \textbf{1.0}   & \textbf{0.071}/\textbf{ 0.105} & 0.752/ 0.762                   & \textbf{0.797}/ \textbf{0.812} & \textbf{0.775}/\textbf{ 0.787}~ 
\end{tblr}
\vspace{-2em}
\end{table*}

\subsection{Multiview Attributes Perceiver}
As shown in \cref{fig: MAP}, after getting the weights from the Multihead Sampler, the final output of MAP is computed through the weighted sum of $O_{VE}$ over $w_{MS}$. Assume $w_{MS}=\{w_1, w_2, ..., w_n\}$ and the output of Visual Extractor $O_{VE} = \{o_1, o_2, ..., o_n\}$, the outputs of MAP can be formulated as follows:
\vspace{-1em}

{\small
\begin{align}
    & \sum \limits_{i=1}^{n} w_i \cdot o_i; \quad o_i \in O_{VE}, w_i \in w_{MS} \notag \\
    & O_{VE} = \{o_1, o_2, ..., o_n\}, w_{MS}=\{w_1, w_2, ..., w_n\} \label{eq:sum}
\end{align}
}
In \cref{eq:sum}, $w_i$ is the corresponding weight of the $i$-th input in $w_{MS}$. Note that in the design of MAP, the number of image inputs is not restricted and this design enables the proposed MIAVLM model to accept any number of image inputs. The form of a weighted sum in the outputs also ensures that the input order has no influence on the final output, making the model more robust and reliable in practice.


\vspace{-1em}
\section{Experiments}
\vspace{-0.5em}

\subsection{Benchmark Settings and Implementation Details}
\vspace{-0.5em}
\textbf{\textit{Benchmark Settings.}} The HoOA benchmark is generated from the CelebAText-HQ \cite{sun2021multi} dataset. In the original dataset, each image contains manually annotated descriptions of facial attributes such as  ear shapes, colors, and various other attributes. Based on these descriptions, we used the Yi-CHAT-34B \cite{young2024yi} model to rewrite them into general questions. 
These questions are called positive questions since all their answers are `Yes'.
To generate negative questions, we use Yi-CHAT-34B \cite{young2024yi} to replace the original attributes in the questions with their opposite words  to generate adversarial question sets. 
Finally, we sampled 1,430 images and obtained 14,291 positive questions and 14,291 negative questions.
During instruction tuning, they were respectively employed as 14,291 positive instructions and 14,266 negative instructions for MIAVLM. Throughout the instruction tuning process, these instructions were divided into training and testing sets in a 9:1 ratio. We define the model's average accuracy on positive and negative questions as the HoOA metric. 

\textbf{\textit{Implementation Details.}} The Language Modeling loss is used for training MIAVLM and we apply Adam optimizer with $lr=0.001$ for optimization. The whole model is trained for 20 epochs with a cosine annealing scheduler. A single NVIDIA 3090 GPU was used for training. 
\vspace{-1em}

\subsection{The Performance of LVLMs on HoOA Benchmark}
\vspace{-0.5em}

We compared MIAVLM (ours) with BLIP3 \cite{xue2024xgen}, four versions of OpenFlamingo \cite{awadalla2023openflamingo}, OPERA \cite{huang2023opera}, Idefics2\cite{laurenccon2024matters} and LLaVA-UHD \cite{xu2024llava} on the HoOA benchmark. Among these LVLMs, both LLaVA-UHD and OPERA claim to have made improvements specifically targeting the hallucination problem based on LLaVA-1.5 \cite{liu2023improved}. All LVLMs utilize two input \textbf{modes}:
\textbf{1}. Using only the original image. \textbf{2}. Using the original image along with eight generated images as input. For LVLMs like BLIP3, LLaVA-UHD, and OPERA, which only support single-image input, \textbf{mode 2} involves combining the nine images into a single image (9in1). 
Overall, we observed that current popular LVLMs generally have a tendency to respond "Yes" to questions. In contrast, our model demonstrates a more balanced approach. Given that our designed \textbf{MAP} can efficiently process multiple images simultaneously and utilizes a lightweight LLM, our model also has a significant advantage in terms of efficiency. By comparing the performance of different LVLMs across the two input modes, we observed that using the 9in1 image as input did not improve results. This may be due to the fact that the 9in1 image is more challenging to interpret compared to the original image. In contrast, models that used nine separate multiview images as input showed overall performance improvements.

To demonstrate the importance of negative instructions, we only used positive instructions to tune MIAVLM. The results are shown in \cref{tab:ni}. It can be observed that using negative instructions effectively enhances the model's performance on negative questions. However, using negative instructions also leads to performance degradation on positive questions.
\vspace{-0.5em}

\begin{table}
\vspace{-1em}
\centering
\caption{Ablation experiments for Negative Instructions (NI).}
\vspace{-1.3em}
\label{tab:ni}
\begin{tblr}{
  width = \linewidth,
  colspec = {Q[331]Q[381]Q[315]},
  cells = {c},
  hlines,
}
                   & MIAVLM (without NI)         & MIAVLM (with NI)               \\
\small{Pos./ Neg./ HoOA} & \textbf{0.790}/ 0.540/ 0.665  & 0.762/ \textbf{0.812}/ \textbf{0.787}
\end{tblr}
\vspace{-2.5em}
\end{table}

\subsection{The Influence of Multiview Images Input Order on LVLMs }
\vspace{-1em}
\begin{figure}[htbp]
\begin{center}
\centerline{\includegraphics[width=0.75\columnwidth]{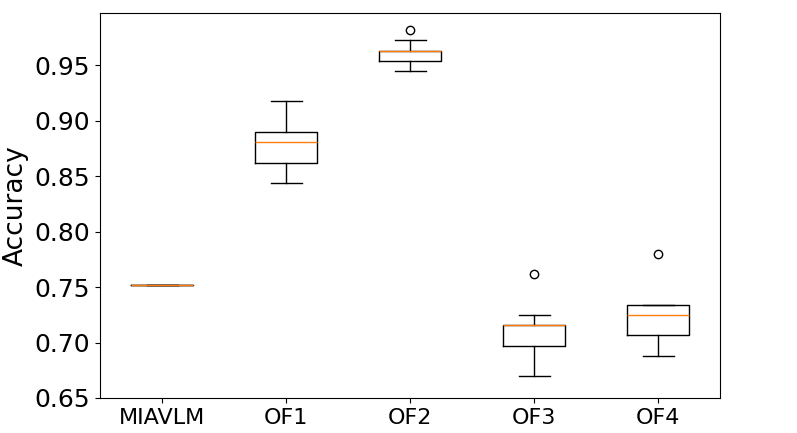}}
\vspace{-1.5em}
\caption{The influence of multiview images input order on OpenFlamingo \cite{awadalla2023openflamingo} and MIAVLM (ours). \Circle 
: Outlier. Yellow line: Median. OF: OpenFlamingo.}
\label{fig: order}
\end{center}
\vspace{-1.5em}
\end{figure}

We used OpenFlamingo \cite{awadalla2023openflamingo} for comparison, along with our MIAVLM, both using 9 images as input from positive questions.  We shuffled the order of these 9 images five times and recorded the results of both models. As shown in \cref{fig: order},  MIAVLM are not affected by any input order. However, any version of OpenFlamingo \cite{awadalla2023openflamingo} is influenced by the order.

\vspace{-0.5em}

\section{Conclusion}
In this paper, we introduce a new benchmark to confirm the significant presence of HoOA problems in popular LVLMs. To mitigate HoOA, we propose MIAVLM, a LVLM that leverages multiview images of the current object as input and employs a novel MAP module to eliminate the influence of input image order. Additionally, negative instructions are utilized to suppress LVLMs' tendency to answer ``Yes" excessively. 


\bibliographystyle{IEEEtran}
\bibliography{IEEEabrv,reference}

\end{document}